\pdfoutput=1

\documentclass{ifacconf}

\usepackage{graphicx}      % include this line if your document contains figures
\usepackage{natbib}        % required for bibliography

\usepackage[english]{babel}

\usepackage[utf8]{inputenc}
\usepackage{amssymb}
\usepackage{amsmath}
\usepackage{mathtools}
\usepackage{commath}
\usepackage{soul} % strike though
\usepackage[sc,osf]{mathpazo}
\usepackage[utf8]{inputenc}

\usepackage{xcolor}

\usepackage{graphicx}

\usepackage{epstopdf}
\graphicspath{{../}}

\usepackage{mwe}

\usepackage[caption=false, font=footnotesize]{subfig}

\usepackage{booktabs}
\usepackage{tabularx}
\usepackage{array}

\usepackage{float}
\floatstyle{plaintop}
\restylefloat{table}

\usepackage{tikz}
\usepackage{pgfplots}
\pgfplotsset{compat=newest}
\usepackage{standalone}
\usepackage{textcomp}
\usepackage{siunitx}

\begin{document}
\begin{frontmatter}

\title{Towards thruster-assisted bipedal locomotion for enhanced efficiency and robustness} 
% Title, preferably not more than 10 words.

% \thanks[footnoteinfo]{Sponsor and financial support acknowledgment
% goes here. Paper titles should be written in uppercase and lowercase
% letters, not all uppercase.}

\author[First]{Pravin Dangol} 
\author[First]{Alireza Ramezani}

\address[First]{Northeastern University, Boston, MA 02115 USA (email: dangol.p@husky.neu.edu, a.ramezani@northeastern.edu)}

\begin{abstract}                % Abstract of not more than 250 words.

In this paper, we will report our efforts in designing closed-loop feedback for the thruster-assisted walking of bipedal robots. We will assume for well-tuned supervisory controllers and will focus on fine-tuning the joints desired trajectories to satisfy the performance being sought. In doing this, we will devise an intermediary filter based on reference governors that guarantees the satisfaction of performance-related constraints. Since these modifications and impact events lead to deviations from the desired periodic orbits, we will guarantee hybrid invariance in a robust way by applying predictive schemes withing a very short time envelope during the gait cycle. To achieve the hybrid invariance, we will leverage the unique features in our model, that is, the thrusters. The merit of our approach is that unlike existing optimization-based nonlinear control methods, satisfying performance-related constraints during the single support phase does not rely on expensive numeric approaches. In addition, the overall structure of the proposed thruster-assisted gait control allows for exploiting performance and robustness enhancing capabilities during specific parts of the gait cycle, which is unusual and not reported before.
\end{abstract}


\end{frontmatter}
%===============================================================================

\maketitle

\section{Introduction}
Raibert's hopping robots \citet{raibert1984experiments} and Boston Dynamic's BigDog \citet{raibert2008bigdog} are amongst the most successful examples of legged robots, as they can hop or trot robustly even in the presence of significant unplanned disturbances. Other than these successful examples, a large number of humanoid robots have also been introduced. Honda's ASIMO \citep{hirose2006honda} and Samsung's Mahru III \citep{kwon2007biped} are capable of walking, running, dancing and
going up and down stairs, and the Yobotics-IHMC \citep{5354430} biped can recover from pushes.

Despite these accomplishments, all of these systems are prone to falling over. Even humans, known for natural and dynamic gaits, whose performance easily outperform that of today's bipedal robot cannot recover from severe pushes or slippage on icy surfaces. Our goal is to enhance the robustness of these systems through a distributed array of thrusters. 

\begin{figure}
    \centering
    \includegraphics[width=0.5\linewidth]{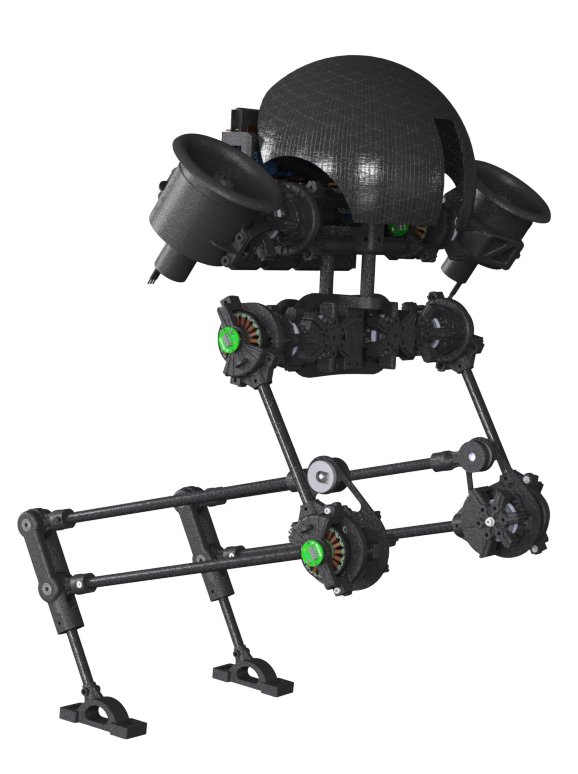}
    \caption{CAD model for a thruster-assisted bipedal robot designed by the authors}
    \label{fig:Leo}
\end{figure}

Here, in this paper, we report our efforts in designing closed-loop feedback for the thruster-assisted walking of legged systems, currently being developed at Northeastern University. These bipeds are equipped with a total of six actuators, and two pairs of coaxial thrusters fixed to their torso. An example is shown in figure \ref{fig:Leo}. 

These platforms combine aerial and legged modality in a single platform and can provide rich and challenging dynamics and control problems. The thrusters add to the array of control inputs in the system (i.e., adds to redundancy and leads to overactuation) which can be beneficial from a practical standpoint and challenging from a feedback design standpoint. Overactuation demands an efficient allocation of control inputs and, on the other hand, can safeguard robustness by providing more resources.

The challenge of simultaneously providing asymptotic stability and constraint satisfaction in legged system has been extensively addressed \citet{westervelt2007feedback}. The method of hybrid zero dynamics (HZD) has provided a rigorous model-based approach to assign attributes such as efficiency of locomotion in an off-line fashion. Other attempts entail optimization-based approaches to secure safety and performance of legged locomotion, see \citet{CLFQP}, \citet{7803333}, and \citet{7041347}.   

Instead of investing on costly optimization-based schemes in single support (SS) phase, we will assume for well-tuned supervisory controllers found in \citet{sontag1983lyapunov}, \citet{371031}, and \citet{bhat1998continuous}. Instead will focus on fine-tuning the joints desired trajectories to satisfy the performance being sought. In doing this, we will devise an intermediary filter based on the emerging idea of reference governors, see \citet{411031}, \citet{bemporad1998reference}, and \citet{gilbert2002nonlinear}. Since these modifications and impact events lead to deviations from the desired periodic orbits, we will guarantee hybrid invariance in a robust fashion by applying predictive schemes withing a very short time envelope during the gait cycle, i.e. double support (DS) phase. To achieve hybrid invariance, we will leverage the unique features in our robot, i.e., the thruster. As a result, the merit of our approach is that unlike existing methods satisfying performance-related constraints during the single support phase does not rely on expensive optimization approaches. In addition, the proposed design approach allows to enhance performance and robustness beyond the limits manifested by existing state-of-the-art dynamic walkers. 

This work is organized as follows. In section \ref{section:method}, the dynamics for a planar three link biped is developed. The SS phase is modeled following standard conventions, then a two-point impact map and a non-instantaneous DS phase are introduced. In SS phase gaits are first designed based on HZD method, constraints are imposed on the states and inputs through an explicit reference governor (ERG). During DS phase a nonlinear model predictive control (NMPC) scheme is utilized to steer states back to zero dynamics manifold ensuring hybrid invariance. Results are shown in section \ref{section:results}, and the paper is concluded in section \ref{section:conclusion}.

\section{Methodology} \label{section:method}
% Methodology section

In the sagittal plane, bipedal locomotion is simplified to an equivalent three-link model. A single gait is divided into two phases including 1) SS phase when only one feet is on the ground and, 2) a DS phase when both feet are grounded. The phases are separated by a discrete transition caused by an impulsive impact force acting on the biped when the swing foot makes contact with the ground. An extended DS phase is considered, unlike widely used assumption of instantaneous DS phase, see \citet{898695}, \citet{westervelt2003hybrid}, \citet{chevallereau2004} \citet{choi_grizzle_2005}, and \citet{1641816}. Here, the DS phase is used to make corrections for error brought on by the impact event. 

% %===================================================================

\subsection{Brief overview of the hybrid model}

\begin{figure}
\begin{minipage}{.46\linewidth}
\centering
    \subfloat[\label{fig:stick:SS}]{
        \includegraphics[width=0.7\linewidth]{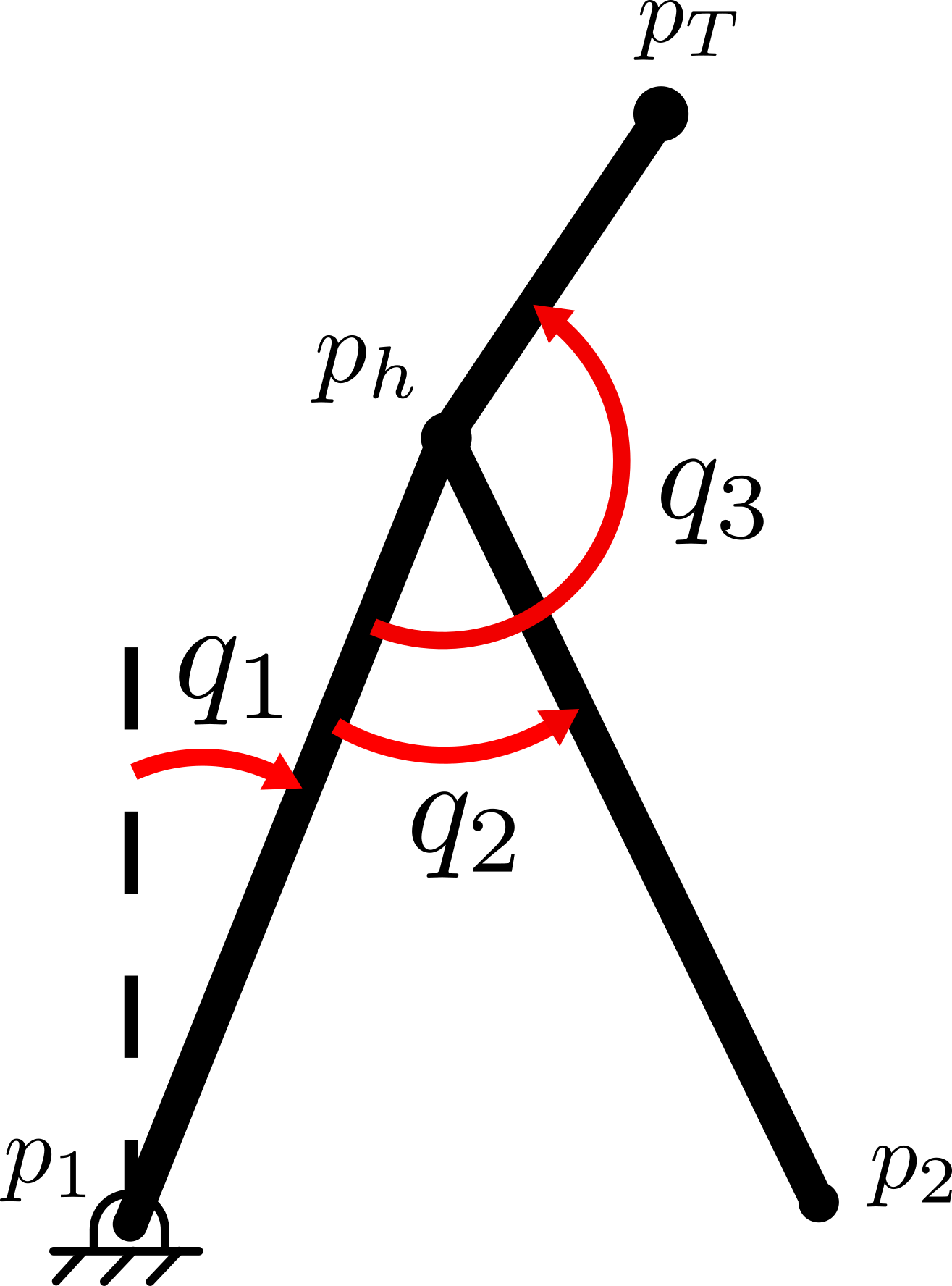} 
        }
\end{minipage}%
% \hfill
\begin{minipage}{.46\linewidth}
\centering
    \subfloat[\label{fig:stick:DS}]{
        \includegraphics[width=0.7\linewidth]{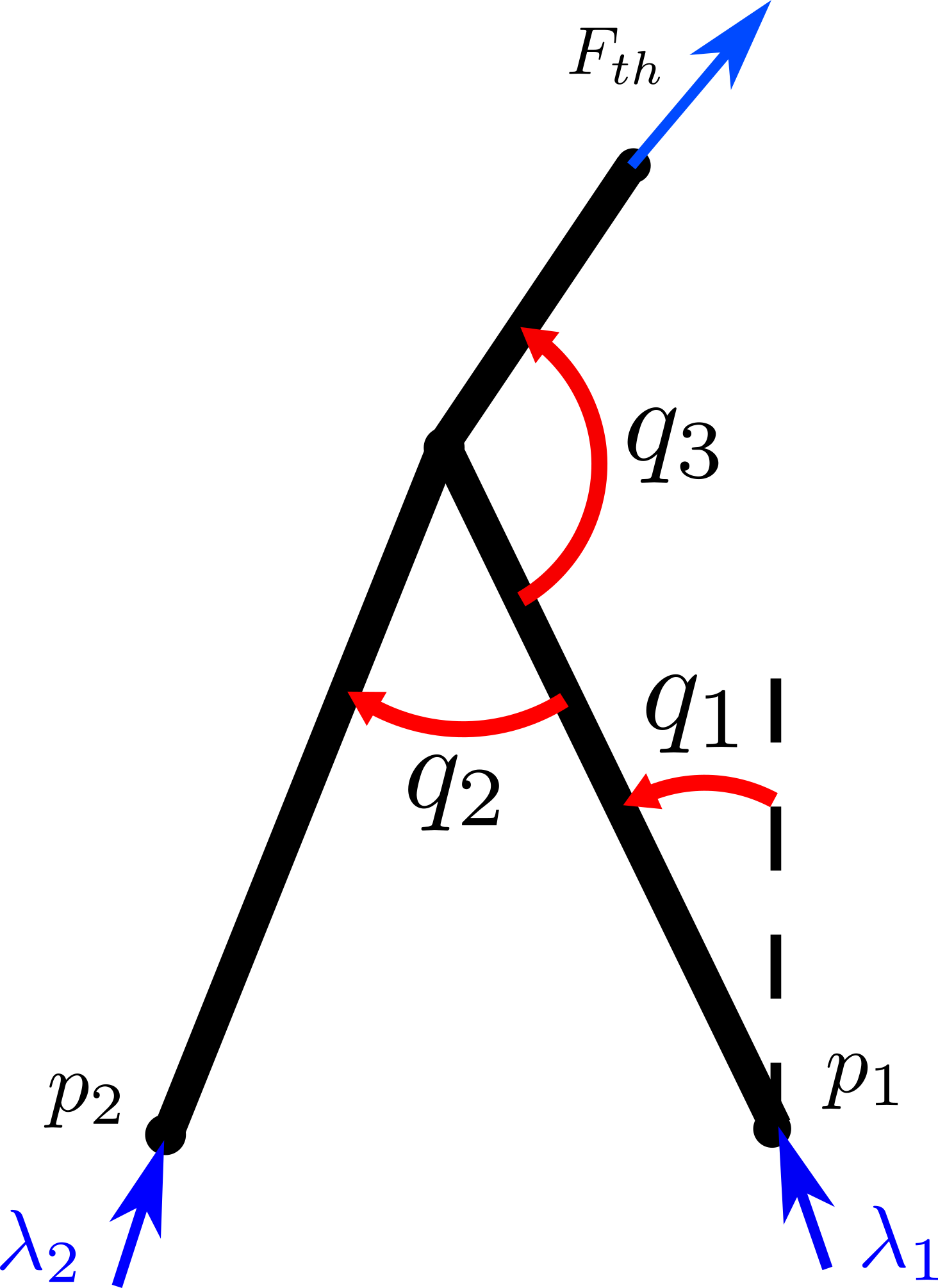}
        }
\end{minipage}
\caption{Stick diagram of pinned model (a), and unpinned (unconstrained) model (b)}
\label{fig:stick}
\end{figure}

During the SS phase the biped has 3 degrees of freedom (DOF) and 2 degrees of actuation (DoA), which yields 1 degree of underactuation (DoU). It is assumed that the stance leg acts as an ideal pivot throughout the phase, i.e., it is fixed to the ground with no slippage. The kinetic $\mathcal{K}(q_s,\dot{q_s})$ and potential $\mathcal{V}(q_s)$ energies are derived to formulate the Lagrangian, $\mathcal{L}(q_s,\dot{q_s}) = \mathcal{K}(q_s,\dot{q_s}) - \mathcal{V}(q_s)$, and form the equation of motion \citet{westervelt2007feedback}:
\begin{equation} \label{eqn:swing}
    D_s(q_s)\Ddot{q_s} + H_s(q_s,\dot{q}_s) = B_s(q_s)u 
\end{equation}

\noindent where $D_s$ is the inertial matrix independent of the underactuated coordinate, $H_s$ matrix contains the Coriolis and gravity terms, and $B_s$ maps the input torques to the generalized coordinates $q_s$. The choice of configuration variables used are as follows: $q_1$ is the absolute stance leg angle which is also the under-actuated coordinate; $q_{2}$ is the angle of the swing leg relative to stance leg; and $q_{3}$ is the angle of torso relative to swing leg as shown in Fig. \ref{fig:stick:SS}. The configuration variable vector is denoted by $q_s = [q_{1}, q_{2}, q_{3}]^T \in \mathcal{Q}_s$. 

% SS and DS stick figures

%===================================================================
The transition between the end of SS phase and the beginning of DS phase is caused by an impulsive impact event when the end of the swing feet $p_2$ makes contact with the ground. This is denoted by a switching surface $\mathcal{S}_s^d = \{x \in \mathcal{TQ} | p_2^V = 0, \dot{p}_2^V < 0\}$. The impact map is modeled as in \citet{impact}, which solves for post impact states and ground reaction forces (GRF). In order to formulate this impact map, the planar model from SS phase is now considered to be unpinned by augmenting $q_s$ to include the hip position, $q_e = [q_s, p_h]^T$. The Lagrangian is reformulated and the impulsive GRF $\delta F_{ext}$ is added on
\begin{equation} \label{eqn:impact}
    D_e(q_e)\Ddot{q_e} + H_e(q_e,\dot{q}_e) = B_e(q_e)u + \delta F_{ext}
\end{equation}

\noindent where $F_{ext}$ acts on the end of each feet $p=[p_1, p_2]^T$ and is expressed as
\begin{equation*}
    F_{ext} = J^T \lambda = \begin{bmatrix} 
    \partial p_1 / \partial q_e \\
    \partial p_2 / \partial q_e \\
    \end{bmatrix}^T 
    \begin{bmatrix} 
    \lambda_1 \\
    \lambda_2 \\
    \end{bmatrix}
\end{equation*}

\noindent here $\lambda$, is a Lagrange multiplier that assumes both legs are fixated to the ground at the moment of impact impact and the Jacobian matrix is given by $J = \partial p(q_e) / \partial q_e$. It is assumed that the impact is inelastic, the angular momentum is conserved and both legs ends are fixed to the ground. 

The last assumption is $\dot{p}^- = J \dot{q}^-_e = 0$, which means that the feet end are fixed to the ground at the time of impact and post impact. Combining this with the conservation of angular momentum allows for the effect of impact to be solved:

\begin{equation}  \label{eqn:impact_matrix_ds}
    \begin{bmatrix}
    \dot{q}_e^+ \\
    \lambda
    \end{bmatrix}    =
    \begin{bmatrix}
    D_e(q_e^-) & -J(q_e^-)^T \\
    J(q_e^-) & 0_{4 \times 4}
    \end{bmatrix} ^{-1}
    \begin{bmatrix}
    D_e(q_e^-)\dot{q}_e^- \\
    0_{4 \times 1}
    \end{bmatrix}
\end{equation}

\noindent where the superscript $+$ denotes post-impact and $-$ denotes pre-impact states. The inertial matrix $D_e$ is square, symmetric and positive definite, and the Jacobian $\partial (p(q_s)+p_h)/\partial q_e$ is always full rank, allowing for the matrix inversion shown on the right hand side.

The impact event also marks the beginning of the next gait, so the roles of legs are swapped post impact yielding
\begin{align}
    \begin{bmatrix}
    q_{1}^+ \\
    q_{2}^+ \\
    q_{3}^+ 
    \end{bmatrix}
    = 
    \begin{bmatrix}
    q_{2}^{-} - q_{1}^{-} \\
    -q_{2}^{-} \\
    q_{3}^{-} - q_{2}^{-}  
    \end{bmatrix}
\end{align}

\noindent which can be captured by a matrix $R_s^d$ as $x_e^+ = [q_{e}^+, \dot{q}_{e}^+]^T = R_s^d[q_e^-, \Delta (\dot{q}_e^-)]^T$. Where $\Delta: q_e^- \mapsto q_e^+$ maps pre-impact to post-impact velocities obtained from (\ref{eqn:impact}).

%===================================================================

\subsection{DS phase with thrusters}
After impact as both feet stay fixed to the ground, this results in a non-instantaneous DS phase, which we assume to occur for a significantly shorter duration than that of the SS phase. The unconstrained dynamics with the ground reaction forces $\lambda$ and the thrusters' action $F_{th}$ as shown in Fig. \ref{fig:stick:DS} are given by 
\begin{equation} \label{eqn:double_support}
    D_{d}(q_{d})\Ddot{q_{d}} + H_{d}(q_{d},\dot{q_{d}}) = B_{d}(q_{d})\eta + J^T \lambda
\end{equation}

\noindent where the control input is augmented to incorporate the effect of thrusters $\eta = [u, F_{th}]^T$, which was inactive during SS phase. The orientation of the thrust vector with respect to the body is assumed to be fixed along the torso link and only changes in the magnitude are considered. A damping term (viscous damping) is considered for numerical stability and ease of integration. The kinematics of leg ends are resolved by 

\begin{equation} \label{eqn:no_acc}
    J \Ddot{q}_{d} + \frac{\partial J}{\partial q_{d}} \dot{q}_{d}^2 + d J \dot{q}_{d}= 0 
\end{equation}
\noindent where $d$ is the damping coefficient. The DS phase dynamical model can then be written as: 
 
\begin{equation} \label{eqn:DS_DAE}
    \begin{bmatrix}
    \Ddot{q}_{d} \\
    \lambda
    \end{bmatrix} = 
    \begin{bmatrix}
    D_{d}(q_{d}) & -J(q_{d})^T\\
    J(q_{d}) &  0_{7 \times 7}
    \end{bmatrix}^{-1}
    \begin{bmatrix}
    B_{d} \eta -  H_{d}(q_{d},\dot{q}_{d}) \\
    -\frac{\partial J(q_{d})}{\partial q_{d}} \dot{q}_{d}^2 - d J \dot{q}_{d}
    \end{bmatrix}
\end{equation}

The end of the DS phase leads to next SS phase and is initiated by the end of swing foot breaking contact with the ground, which is defined as $\mathcal{S}_d^s = \{x \in \mathcal{TQ} | p_2^V > 0, \dot{p}_2^V > 0\}$. The initial SS states are simply the final DS states when the swing leg lifts off.

%===================================================================

\subsection{Motion control}
The trajectories for the actuated coordinates are designed by imposing virtual constraints as in \citet{westervelt2007feedback}. The restricted dynamics $f_z=f(x_s)+g(x_s)u^*$ on the zero dynamics manifold $\mathcal{Z}$ are prescribed by the feedback linearizing controller $u^*(x) = - L_g L_f h(x)^{-1} (L^2_f h(x))$ and are invariant of the SS dynamics. This idea is key to HZD-based motion design widely applied to gait design and closed-loop motion control by enforcing holonomic constraint $y = h(x) = q_a - h_d \circ \theta(q)=0$. Where, $q_a = [q_{2}, q_{3}]^T$ is the vector of actuated coordinates, and $h_d$ is parametrized over the zero dynamics state $\theta(q)$. During the SS phase, the HZD method was used to obtain desired trajectories for $q_a$ and an ERG-based framework was then used to respect limits on inputs; during DS phase a NMPC scheme is used to ensure impact invariance by leveraging the thrusters.

%===================================================================

\subsection{SS phase control}

To ensure the actuated coordinates follow the virtual constraints, a variety of finite time convergence controller can be utilized. In our case, with a relative-degree 2 the feedback linearizing control law is $u = -L_g L_f h(x)^{-1} (L_f^2 h(x) + v)$ as in \citet{khalil2002nonlinear}, where $v = K_P y + K_D \dot{y}$ is one of the simplest form of controllers available. In order to ensure that the physical limits on states and inputs are satisfied the idea of reference governor described in \citet{gilbert2002nonlinear} is taken. However, their work involves optimization and to avoid that we took an optimization-free approach based on explicit reference governor (ERG) idea described in \citet{garone2015explicit}.

The ERG acts as a supervisory controller which in our case will manipulate velocity trajectories The main idea here is that constraints on inputs and states can be satisfied by adding dynamics to the reference trajectories rather than assuming them to be pre-defined. This changes the original output functions to:
\begin{align}\label{eqn:y}
\begin{split}
    \dot{y} &= \dot{q}_a - w 
\end{split}
\end{align}

\noindent where $w$ is the manipulated reference that estimates $\dot{h}_d$ while ensuring constraint satisfaction described below. The relative degree of 2 is still preserved even with this change.

\noindent An approach based on Lyapunov argument is taken to formulate the manipulated reference dynamics $\dot{w}$. This is achieved through setting an upper bound on a Lyapunov function of the actuated coordinates such that state and control limits specified in the vector $C(x_a,w)$ are always satisfied. This vector is defined as following
\begin{align}\label{eqn:c_ERG}
    C(x_a,w) &:= C_{x} x_{a} + C_{w} x_w + C_{limit} \geq 0
\end{align}
\noindent where $x_{a} = [q_a, \dot{q}_a]^T$, $x_w = [0, w]^T$ and $C_{x}$, $\ C_{w}$ and $\ C_{limit}$ arise from the limits applied to the states and inputs, that is, $|x| \leq x_{max}$ and $|u| \leq u_{max}$. These limits can be expanded as following:
\begin{align}
\begin{split}
    -q_{max} \leq q &\leq q_{max} \\
    -\dot{q}_{max} \leq \dot{q} &\leq \dot{q}_{max} \\
    -L_g L_f h^{-1}(L_f^2 h + v) &\leq u_{max} \\
    -L_g L_f h^{-1}(L_f^2 h + v) &\geq -u_{max} \\
\end{split}
\end{align}

\noindent where, $v = K_P (q - h_d) + K_D (\dot{q} - w)$. These inequalities can then be rearranged to fit the form in (\ref{eqn:c_ERG}) as following:

\begin{align}
    \begin{split}
    & C_x = 
        \begin{bmatrix}
        I_{4 \times 4} \\
        -I_{4 \times 4} \\
            \begin{matrix}
            K_{P} & K_{D} \\
            -K_{P} & -K_{D} 
            \end{matrix}
        \end{bmatrix} 
     \quad C_w = 
        \begin{bmatrix}
        0_{8 \times 4} \\
            \begin{matrix}
            0 & K_{D} \\
            0 & -K_{D} 
            \end{matrix}
        \end{bmatrix}
    \\
    & C_{limit} = 
        \begin{bmatrix}
        x_{max} \\
        x_{max} \\
        L_g L_f h(x) u_{max} - L^2_f h(x) + K_P h_d\\
        L_g L_f h(x) u_{max} - L^2_f h(x) + K_P h_d \\
        \end{bmatrix} 
    \end{split}
\end{align}

The following Lyapunov function $V(x_{a}, x_w)$ is considered 
\begin{align}
    V(x_{a},x_w) = (x_{a} - x_w)^T P (x_{a} - x_w)
\end{align}
\noindent where $P$ is a positive definite matrix consisting of controller gains $K_P$ and $K_D$ ($P = \frac{1}{2} diag(K_p, K_d) >0$). The dynamics of the manipulated reference is defined such that the Lyapunov function is bounded by a smooth positive definite function $\Gamma(w)$, as following 
\begin{align} \label{eqn:V}
    V(x_{a},x_w) \leq \Gamma (w)
\end{align}

Through a change of coordinates $\Tilde{x} = P^{1/2}(x_a - x_w)$, (\ref{eqn:c_ERG}) takes the form $C_{x} P^{-1/2}\Tilde{x} + C_{x}x_w+ C_{w} x_w + C_{limit} \geq 0$. By solving for $\Tilde{x}$ at the boundary of the constraint $C_(x_a,w) = 0$, we obtain 
\begin{align}
    \Tilde{x} = -\frac{P^{-1/2}C_{x}^T}{C_{x} P^{-1} C_{x}^T}(C_{x}x_w+ C_{w}x_w + C_{limit})
\end{align}
\noindent Then the upper bound can be defined as $\Gamma(w) = \Tilde{x}^T \Tilde{x}$, which is the distance from $x_w$ to the boundary of $C(x_{a},x_w)$. 

The time derivative of the manipulated reference $w$ given by  
\begin{align} \label{eqn:dot_w}
    \dot{w} &:= \kappa (\Gamma(w) - V(x_a, w)) sign(\dot{h}_d - w) 
\end{align}
\noindent yields $\dot{\Gamma} (w, \dot{w}) \leq 0$ where $\kappa > 0$ is an arbitrary large scalar. This choice ensures that an attractive vector field is generated pointing towards $\dot{h}_d$. Looking at the time derivative of (\ref{eqn:V}), which yields the following
\begin{align}\label{eqn:V_dot}
    \dot{V}(x_{a},w,\dot{w}) \leq \dot{\Gamma} (w, \dot{w})
\end{align}

\noindent we see that $\dot{V}(x_{a},w,\dot{w})$ is negative semi-definite. Therefore, asymptotic convergence to $\dot{h}_d$ must be verified through LaSalle's principle by showing $V(x_a, w)=0$ holds true only for a finite time. The time derivative of $\Gamma (w)$ is found to be 
\begin{equation}
    \dot{\Gamma} (w, \dot{w}) = 2 \Tilde{x}^T \frac{P^{-1/2}C_{x}^T}{C_x P^{-1} C_x^T} (C_x + C_w)\dot{x}_w
\end{equation}

\noindent From (\ref{eqn:dot_w}) and (\ref{eqn:V_dot}) $\dot{V}(x_{a},w,\dot{w}) = 0$ is only possible when $\dot{w} = 0$.  This happens if $w = \dot{h}_d$, i.e. when convergence is achieved, or when $\Gamma(w) = V(x_a, w)$. In the latter case, when $\dot{w} = 0$, $\Gamma(w)$ as well as $w$ remain constant. For a constant reference $V(x_a, w)$ will decrease after a finite time and convergence is resumed. 

The controller must then be altered to account for this change:
\begin{equation}
    u = \beta_1 (\dot{w} + v) + \beta_2 
\end{equation}

\noindent where $\beta_1 = -D_3 D_1^{-1} D_2 + D_4$ and $\beta_2 = -D_3 D_1^{-1} H_1 + H_2$ are obtained by partitioning the dynamics in (\ref{eqn:swing}) and solving for $\ddot{q}_b$ in $\ddot{y}$. The dynamics are partitioned as follows:
\begin{equation}
    \begin{bmatrix}
    D_1 & D_2 \\
    D_3 & D_4
    \end{bmatrix}
    \begin{bmatrix}
    \ddot{q}_1 \\
    \ddot{q}_a
    \end{bmatrix}
    +
    \begin{bmatrix}
    H_1 \\
    H_2
    \end{bmatrix}
    =
    \begin{bmatrix}
    0 \\
    u
    \end{bmatrix}
\end{equation}

\subsection{Impact invariance}
The two-point impact renders all joints except the torso to be fixed, causing a large deviation in velocities from the reference trajectory and subsequently deviation from the zero-dynamics manifold as well. 
For periodic gaits to be achieved, the SS phase dynamics must be invariant to such deviations. Since the joint actuators are not able to make corrections needed to steer the states back to the zero dynamics manifold ($\mathcal{Z}$), the thrusters are now leveraged in the DS phase to achieve hybrid invariance. Impact invariance such that $\Pi(\Delta(\mathcal{S} \cap \mathcal{Z})) \subset \mathcal{Z}$ is sought, where $\Pi: x_{d,0} \mapsto x_{d,f}$ maps the initial states of DS phase $x_{d,0}$ to initial states of SS phase $x_{s,0}$. With this condition satisfied hybrid invariance will ensure that each gait starts with the same initial condition despite the impulsive effects of impact and deviation from designed trajectories. When DS phase is absent, hybrid invariance takes the from $\Delta(\mathcal{S} \cap \mathcal{Z}) \subset Z$ as in \citet{westervelt2003hybrid}. 

As opposed to the SS phase, the constraints in the DS phase take a more complex form where the ground reaction forces need to be satisfied, $x_{d,f}$ must match the initial states at the SS phase ($x_{s,0}$) to ensure hybrid invariance. We apply a NMPC-based design scheme to steer the post-DS states back to the zero-dynamics manifold. This scheme is known for being costly, however, the duration of the DS phase is significantly shorter than SS. 

Note that a reference for each DS state $r_d[k]$ is generated at every k-th sample over the duration of the double support phase. The reference can be a simple linear trajectory between the post-impact states $x_{d,0}$ and the initial SS phase state $x_{s,0}$. 

The continuous DS phase model in state space form is given by $\dot{x}_d= f(x_{d}) + g(x_{d})\eta$, this model is discretized and linearized at each each sample time. The following optimization problem is then solved, by minimizing the cost function $\phi (x_d,\eta)$:
\begin{align} \label{eqn:cost}
    \begin{split}
    \quad & \underset{\eta[k]}{\text{min}}~  \phi (x_d,\eta)   = \sum_{k=1}^{N}  \sum_{i=1}^{p} w_{x,i} ( x_{d,i}[k] - r_{d,i}[k] ) + \\ & \qquad \qquad \qquad \ \ \sum_{k=1}^{N-1} \sum_{j=1}^{m} w_{\eta,j} \Delta \eta_j[k] \\
    & \text{subj. to:} \\ 
    & x_{d}[1] = R_s^d x_e^+\\
    & x_{d}[k+1] = f(x_d[k]) + g(x_d[k])\eta[k] + A(x_d[k], \eta[k])\\
    & | \eta[k] | < \eta_{max} \\
    & | x_{d}[k] | < x_{d \ max} \\
    & \lvert \lambda_T[k] \rvert  < \mu_s \lvert \lambda_N[k] \rvert \\
    & \lambda_N[k] > 0\\
    \end{split}
\end{align}
\noindent where the initial state of DS phase $x_{d}[1]$ comes directly from the post impact state $x_e^+$, after the roles of the legs have been swapped which is denoted by $R_s^d$ matrix. The subsequent constraint $x_d[k+1]$ ensures that the discrete linearized states belong to the DS phase, where the $A$ matrix contains the linear terms of $\dot{x}_d$ from (\ref{eqn:double_support}). Limits are imposed on both states and control actions through $\eta_{max}$ and $x_{d, \ max}$, respectively. And finally, the ground contact condition must be satisfied for the DS phase i.e., the ratio of tangential $\lambda_T$  to normal forces $\lambda_N$ is less than the static coefficient of friction $\mu_s$ and normal force is always positive. 

With these constrained satisfied, the NMPC guides the DS states towards the initial condition of SS phase, resulting in impact and DS phase invariance.

%===================================================================

\section{Results} \label{section:results}

\begin{figure}
\begin{minipage}{.32\linewidth}
\centering
    \subfloat[\label{fig:joint:angle1}]{
        \includegraphics[width=1\linewidth]{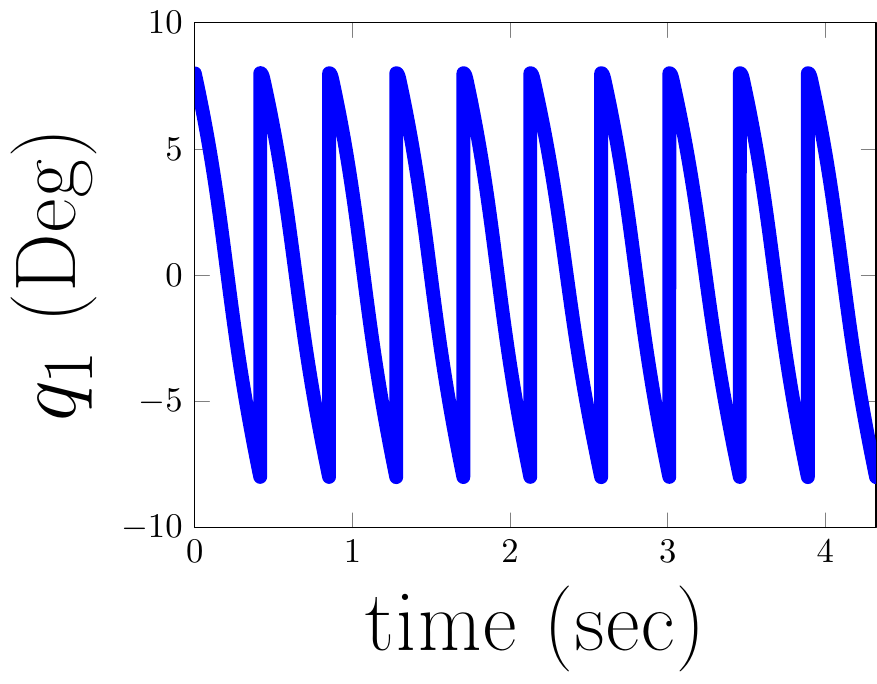} 
        }
\end{minipage}%
\begin{minipage}{.32\linewidth}
\centering
    \subfloat[\label{fig:joint:angle2}]{
        \includegraphics[width=1\linewidth]{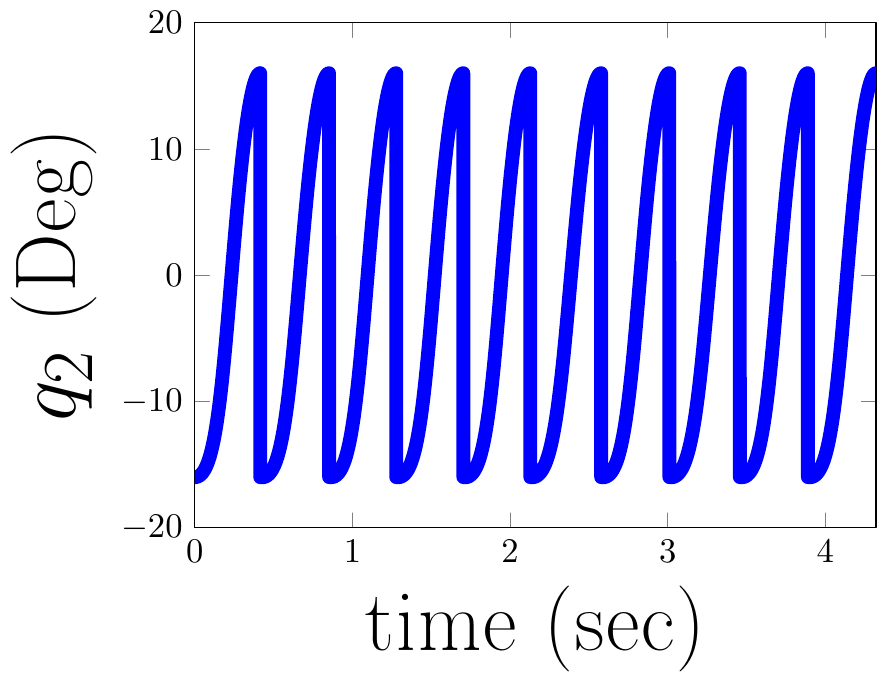} 
        }
\end{minipage}%
\begin{minipage}{.32\linewidth}
\centering
    \subfloat[\label{fig:joint:abgle3}]{
        \includegraphics[width=1\linewidth]{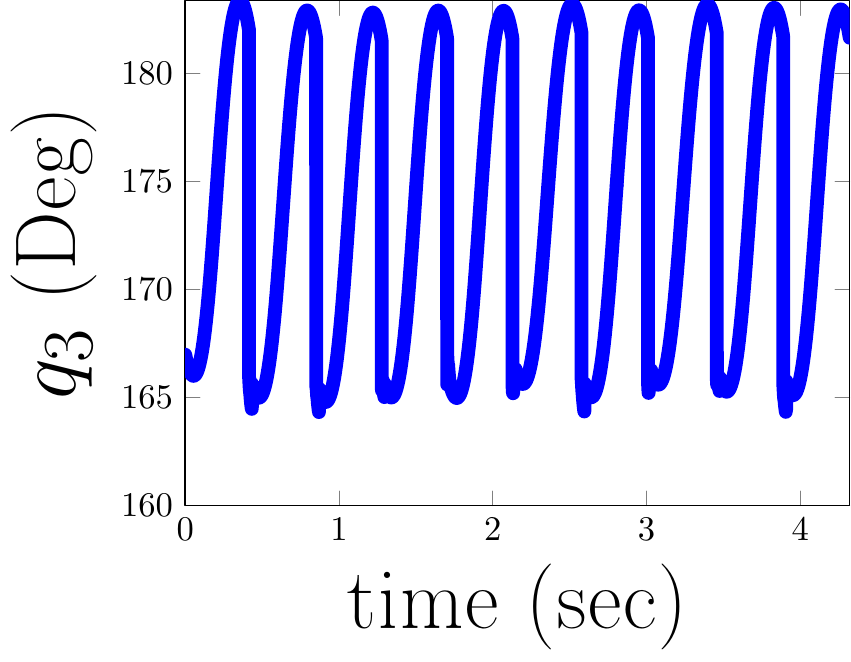}
        }
\end{minipage} \par\medskip
\begin{minipage}{.32\linewidth}
\centering
    \subfloat[\label{fig:joint:vel1}]{
        \includegraphics[width=1\linewidth]{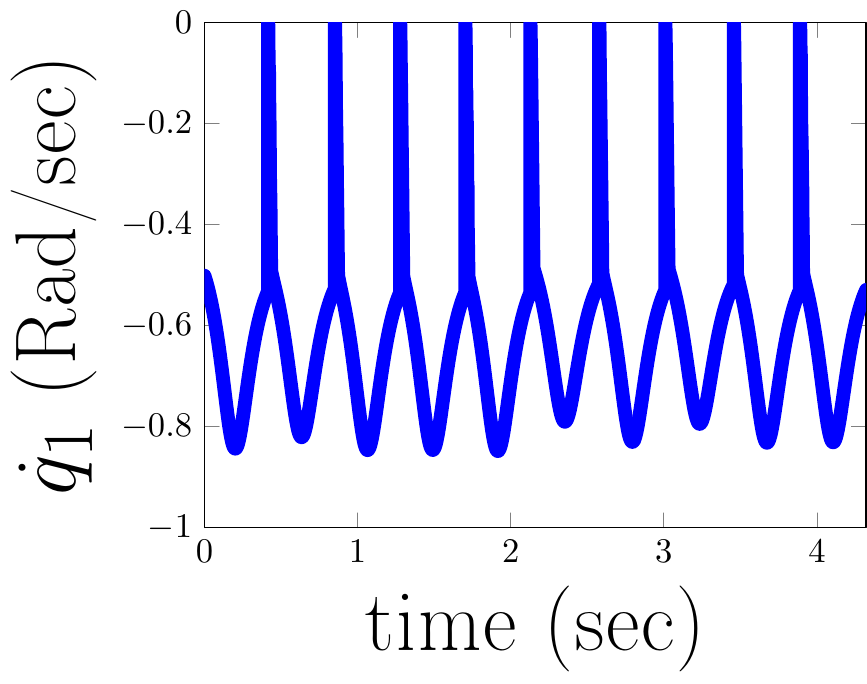} 
        }
\end{minipage}%
\begin{minipage}{.32\linewidth}
\centering
    \subfloat[\label{fig:joint:vel2}]{
        \includegraphics[width=1\linewidth]{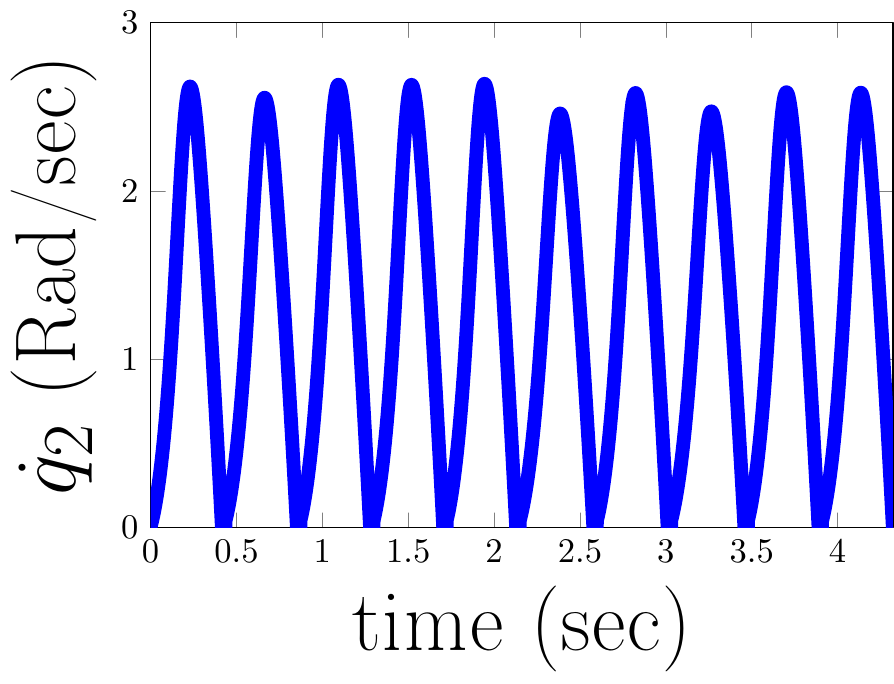} 
        }
\end{minipage}%
\begin{minipage}{.32\linewidth}
\centering
    \subfloat[\label{fig:joint:vel3}]{
        \includegraphics[width=1\linewidth]{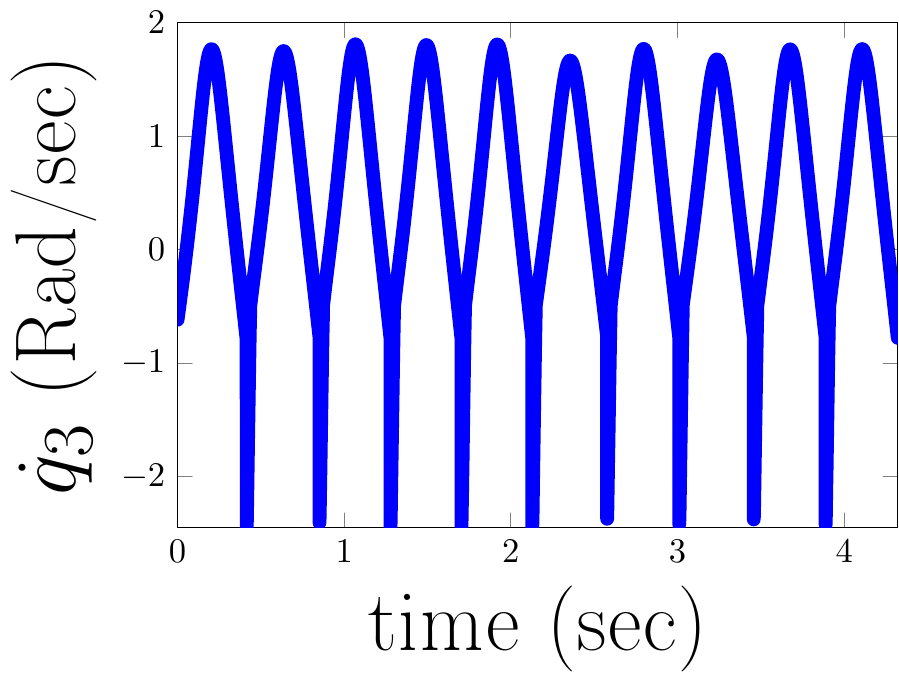} 
        }
\end{minipage}%
\caption{Configuration angle (top row) and velocity (bottom) trajectories of the biped walking 10 steps}
\label{fig:joint}
\end{figure}

For the three link model developed in section \ref{section:method}, a total of 10 steps were simulated. Each DS phase was simulated for a fixed time envelope of 20 mili-seconds. A list of all model parameters used are shown in table \ref{tab:modelparam}. The desired trajectories $h_d$ were generated offline. Figure \ref{fig:joint} shows the configuration variable evolution. The angles ($q_{1,2,3}$) are shown in the first row and the angular velocities ($\dot{q}_{1,2,3}$) are shown in the second row. 

\begin{table} 
\begin{center}
    \begin{tabular}{l r l l} 
        %\\ \toprule
        Parameter &  Value & Description &  \\
        \hline
        $m_T$ &  300.00 $g$ & Mass of torso &  \\
        $m_h$ &  200.00 $g$ & Mass of hip &  \\
        $m_k$ &  100.00 $g$ & Mass of each leg &  \\
        $l_T$ &  30.00 $cm$ & Length from hip to torso &  \\
        $l$ &  63.25 $cm$ & Length of each leg &  \\
        \hline
    \end{tabular}
\end{center}
    \caption{Model Parameters}
    \label{tab:modelparam}
\end{table} 

\vspace{1em}

\begin{figure}
\begin{minipage}{.45\linewidth}
\centering
    \subfloat[\label{fig:DS:mu1}]{
        \includegraphics[width=1\linewidth]{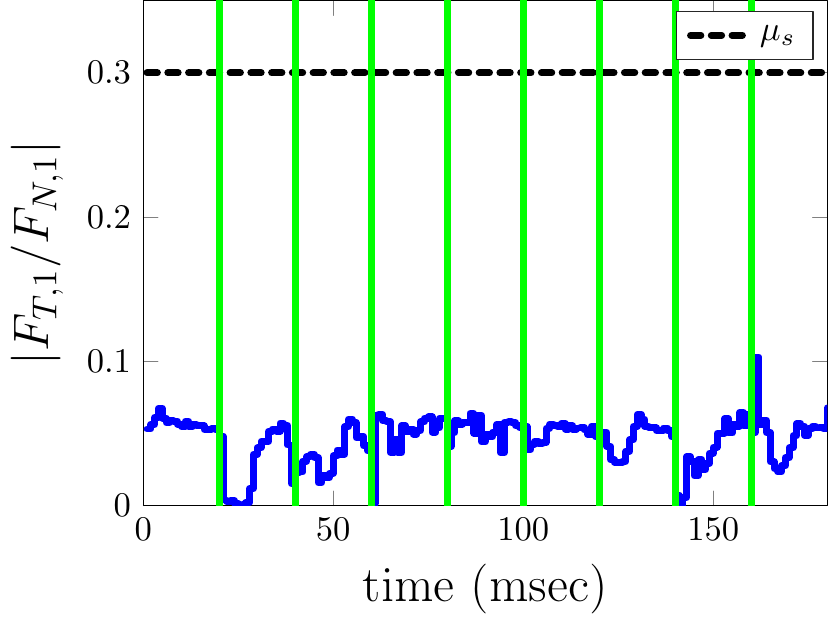} 
        }
\end{minipage}%
\begin{minipage}{.45\linewidth}
\centering
    \subfloat[\label{fig:DS:N1}]{
        \includegraphics[width=1\linewidth]{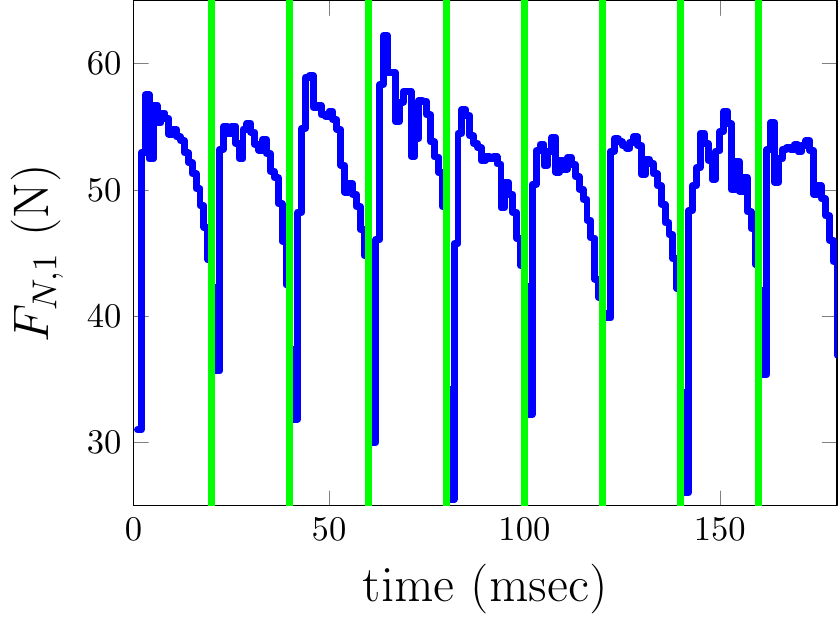} 
        }
\end{minipage} \par\medskip
\begin{minipage}{.45\linewidth}
\centering
    \subfloat[\label{fig:DS:mu2}]{
        \includegraphics[width=1\linewidth]{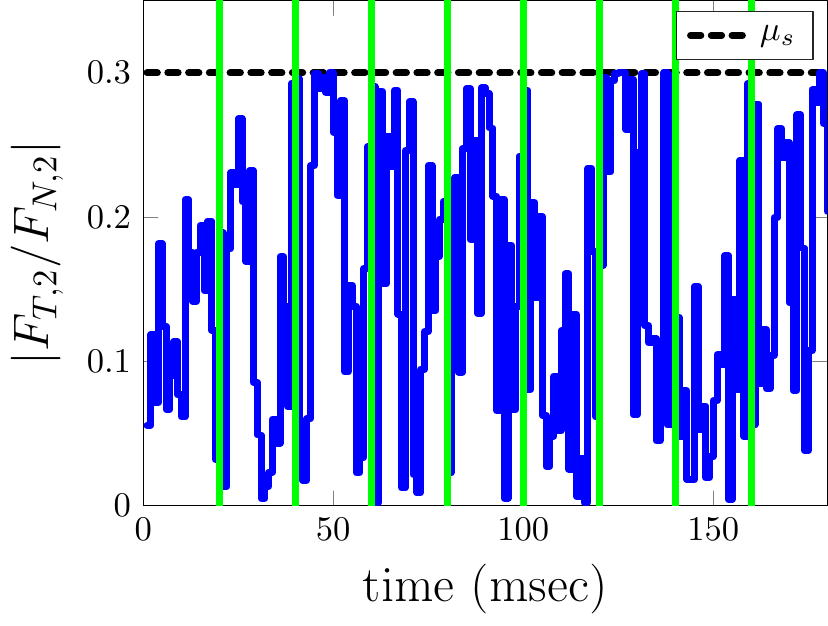} 
        }
\end{minipage}%
\begin{minipage}{.45\linewidth}
\centering
    \subfloat[\label{fig:DS:N2}]{
        \includegraphics[width=1\linewidth]{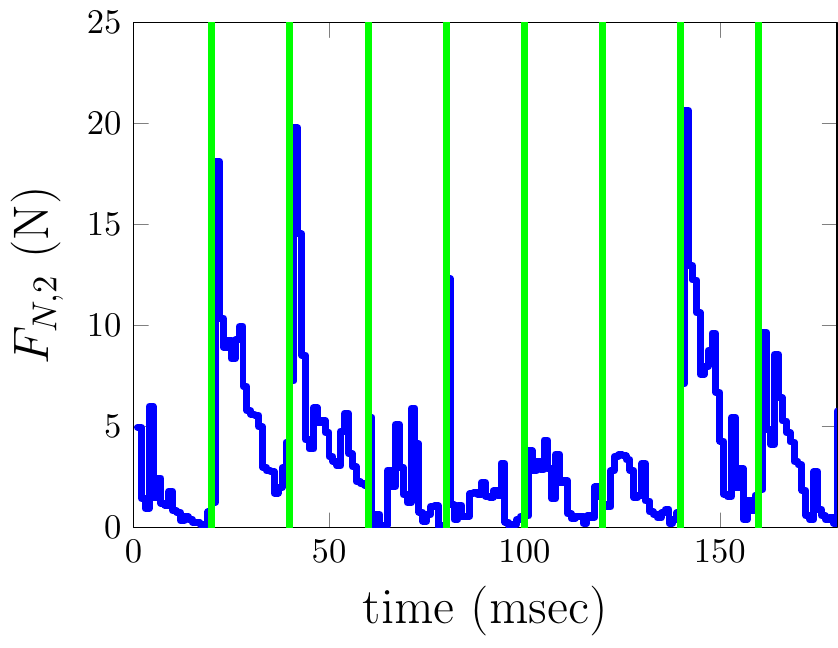} 
        }
\end{minipage}%
\caption{Ground contact force conditions at leg end 1 (a,b) and leg end 2 (c,d), intermediate SS phases are omitted. The vertical green lines indicate consecutive DS phases.}
\label{fig:DS}
\end{figure}

\begin{figure}
    \centering
    \includegraphics[width=0.6\linewidth]{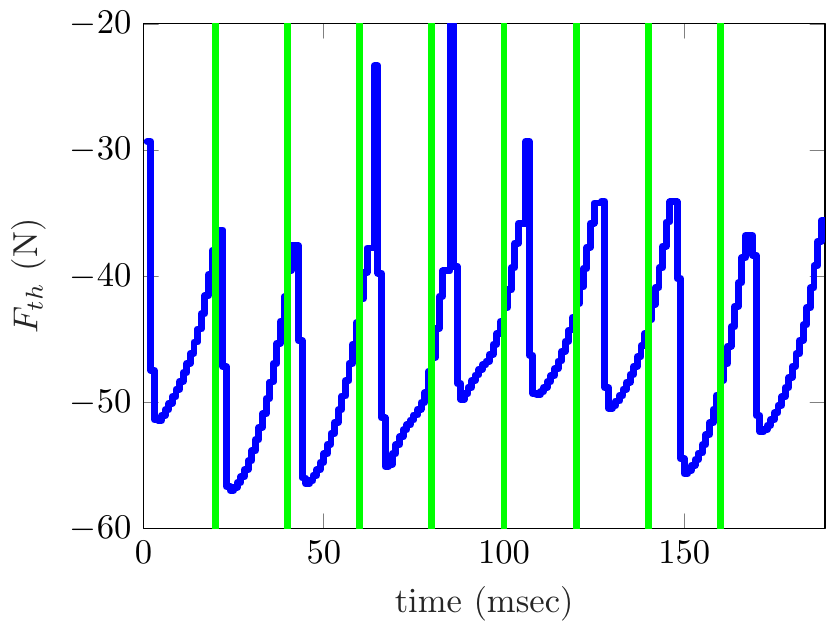}
    \caption{Thruster action during the DS phase}
    \label{fig:DS_Fth}
\end{figure}

Figure \ref{fig:DS} shows that the feasibility conditions are satisfied during the DS phase during the robustification process. The tangential to normal load ratio for each feet is less than or equal to the friction constant value $\mu_s$ at all times and the normal forces are always positive, which indicates that the feet were stuck to the ground throughout the DS phase. We note that the static coefficient of friction is assumed to be $\mu_s = 0.3$ for this simulation study. We also note that that the normal forces spiked to about 60 N and this unusually behavior would not be possible without the inclusion of thrusters' action in the DS phase as the total weight of the biped is only 0.7 kg and the inertial force contributions cannot be directly applied to regulated the ground contact forces. Figure \ref{fig:DS_Fth} shows the control actions for the thrusters during the DS phase. In Fig.~\ref{fig:DS_Fth}, the intermediate SS phases are omitted and the green vertical lines separate consecutive DS phases at each gait cycle.

\begin{figure*}
\centering
% u_max = 30
\begin{minipage}{.3\linewidth}
\centering
    \subfloat[\label{fig:ERG:u2_30}]{
        \includegraphics[width=1\linewidth]{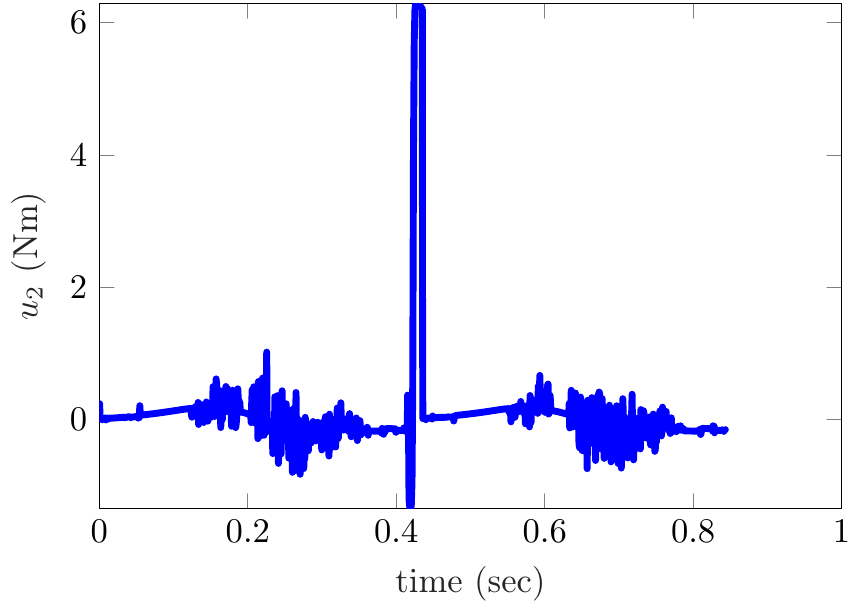} 
        }
\end{minipage}%
\begin{minipage}{.3\linewidth}
\centering
    \subfloat[\label{fig:ERG:u3_30}]{
        \includegraphics[width=1\linewidth]{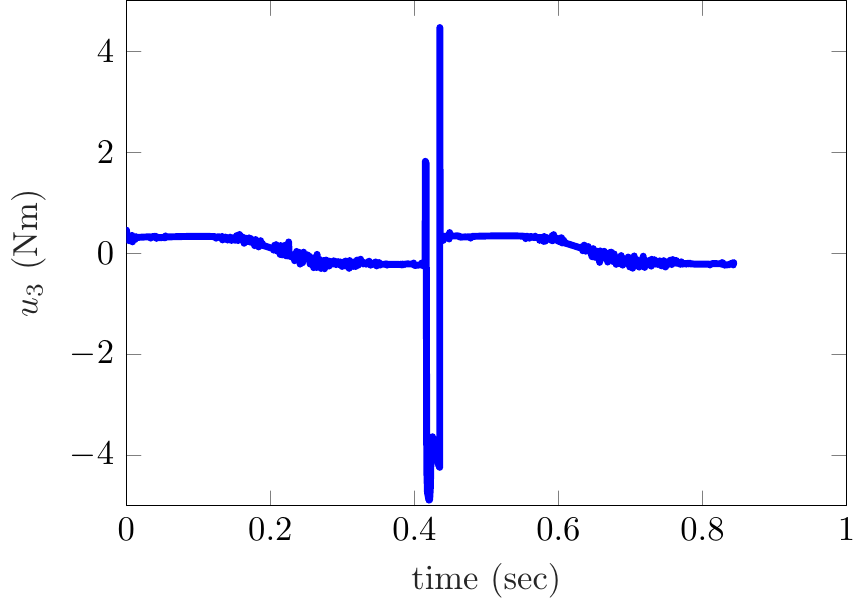}
        }
\end{minipage}
\begin{minipage}{.3\linewidth}
\centering
    \subfloat[\label{fig:ERG:pp_30}]{
        \includegraphics[width=1\linewidth]{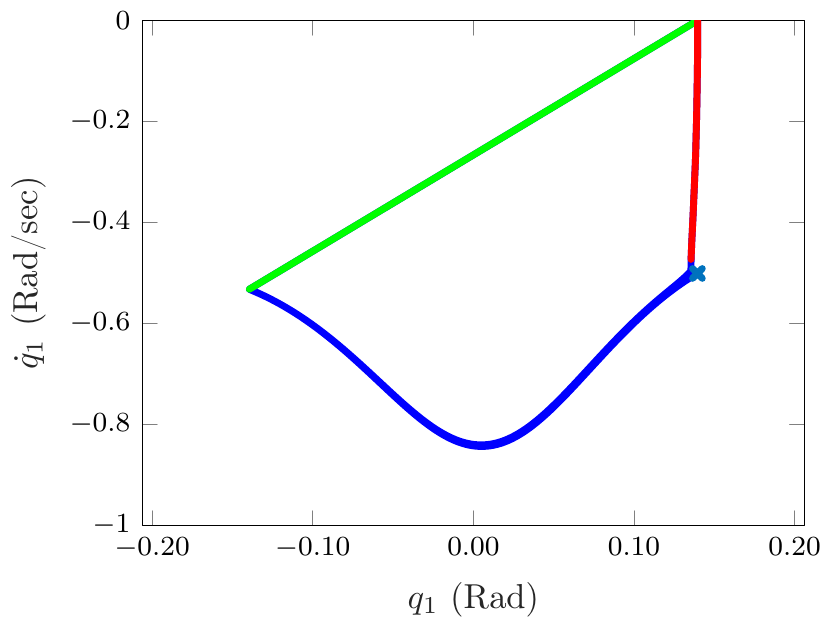}
        }
\end{minipage}\par\medskip
%
% u_max = 15
\begin{minipage}{.3\linewidth}
\centering
    \subfloat[\label{fig:ERG:u2_15}]{
        \includegraphics[width=1\linewidth]{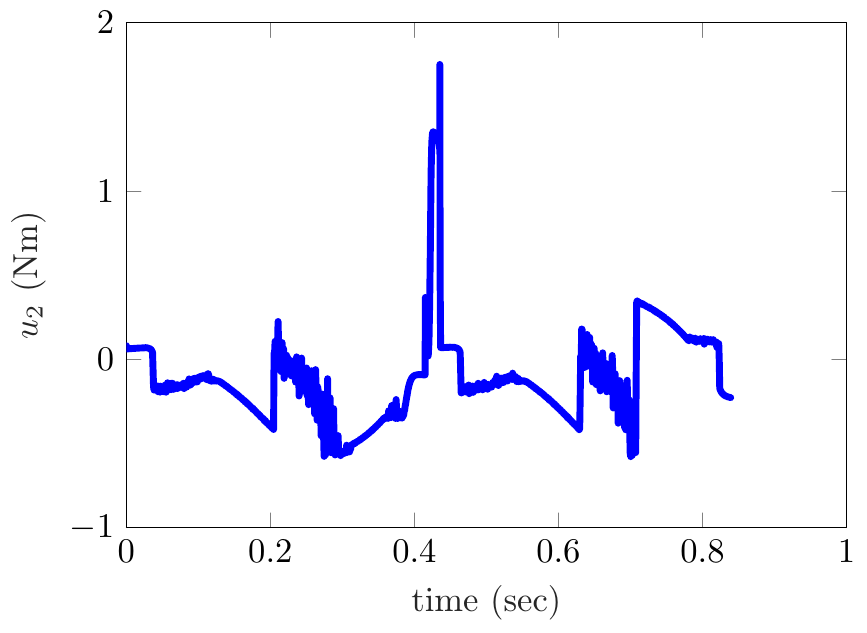} 
        }
\end{minipage}%
\begin{minipage}{.3\linewidth}
\centering
    \subfloat[\label{fig:ERG:u3_15}]{
        \includegraphics[width=1\linewidth]{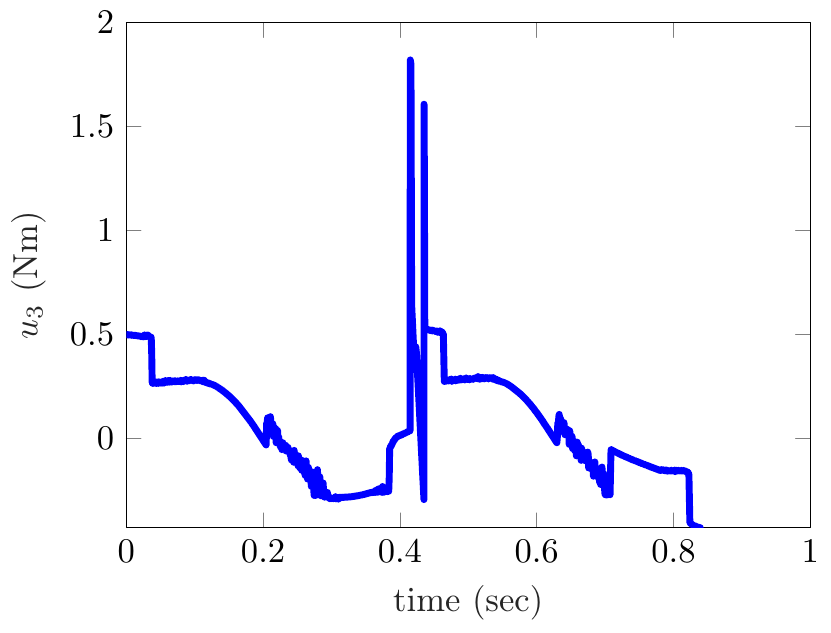}
        }
\end{minipage}
\begin{minipage}{.32\linewidth}
\centering
    \subfloat[\label{fig:ERG:pp_15}]{
        \includegraphics[width=1\linewidth]{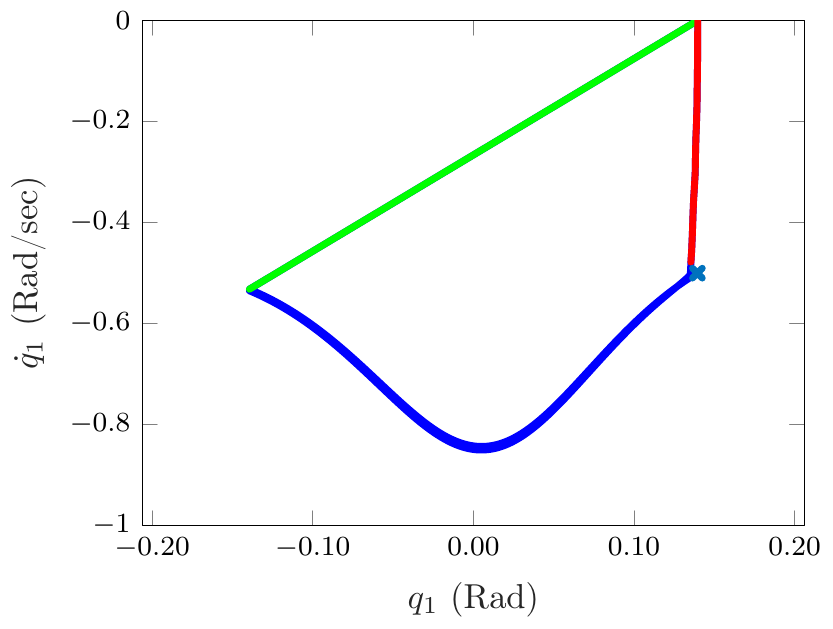}
        }
\end{minipage}
%
% u_max = 10
\begin{minipage}{.3\linewidth}
\centering
    \subfloat[\label{fig:ERG:u2_10}]{
        \includegraphics[width=1\linewidth]{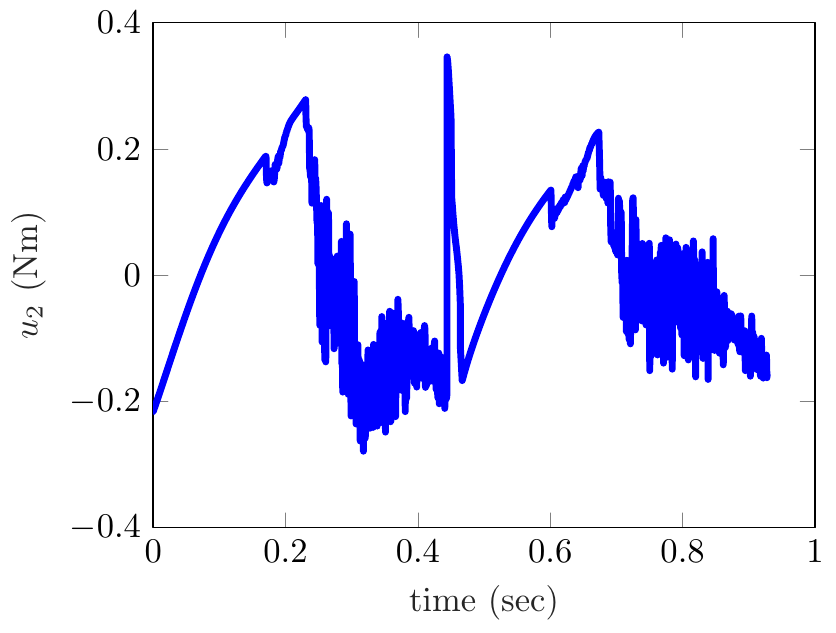} 
        }
\end{minipage}%
\begin{minipage}{.3\linewidth}
\centering
    \subfloat[\label{fig:ERG:u3_10}]{
        \includegraphics[width=1\linewidth]{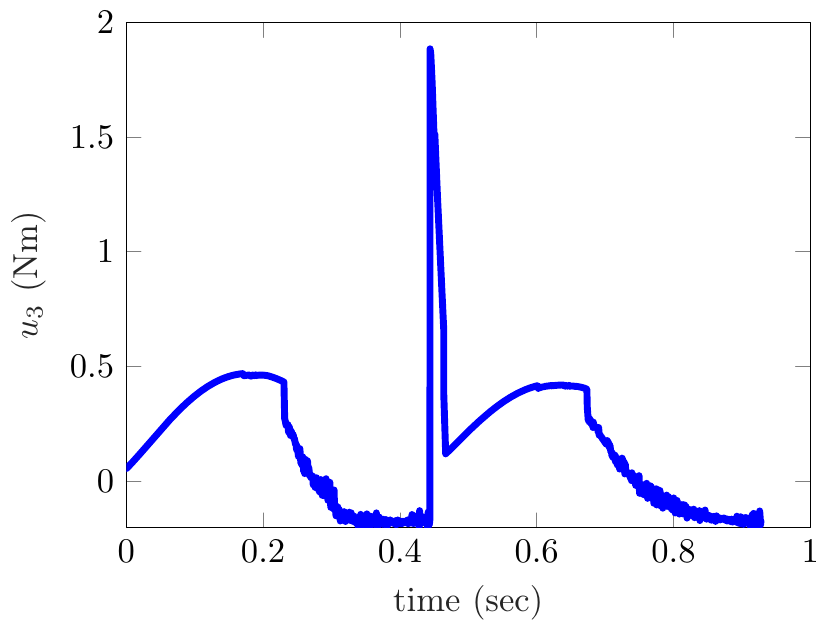}
        }
\end{minipage}
\begin{minipage}{.3\linewidth}
\centering
    \subfloat[\label{fig:ERG:pp_10}]{
        \includegraphics[width=1\linewidth]{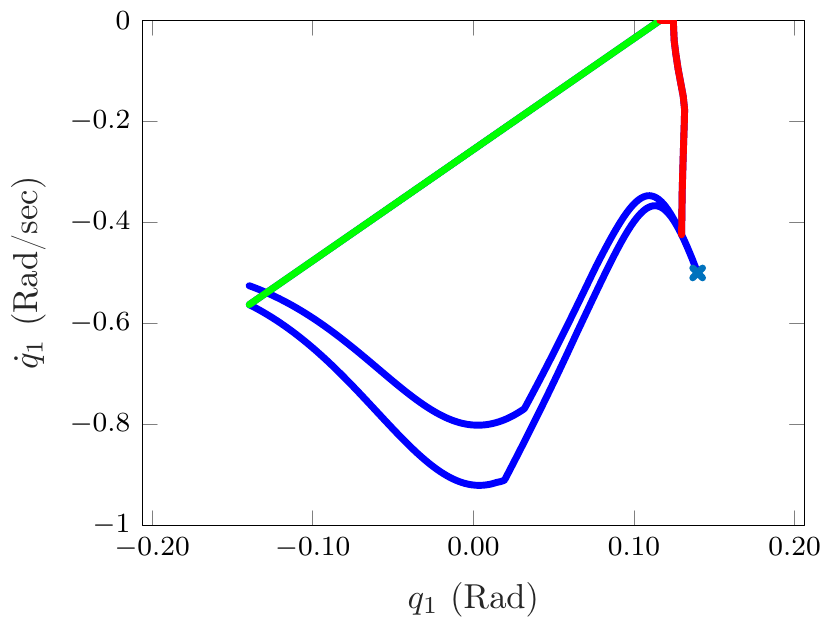}
        }
\end{minipage}
\caption{ERG and NMPC performance; (a,b,c) show control actions and the corresponding phase portrait for the underactuated coordinate q1 when a generous limit is applied on inputs; (d,e,f) and (g,h,i) show the inputs and tracking performance for moderate and conservative limits on inputs, respectively.}
\label{fig:ERG}
\end{figure*}

The synergistic thruster and joint action gait stabilization is summarized in Fig.~\ref{fig:ERG}. The first row shows a generous limit on the joint control actions during the SS phase whereas the third row assumes for a conservative limitation. The phase portrait for the under-actuated coordinate $q_1$ corresponding to these two scenarios are compared in Fig.~\ref{fig:ERG}, where the SS, DS phase and the impact are in blue, red and green, respectively. In the case where the control actions are saturated at a higher value the states converge to the desired limit cycle and as the saturation limits are reduced the tracking performance degrades and the trajectories deviate from the limit cycle to satisfy the constraint. This can raise hybrid invariance issues and during the DS phase this issue is addressed as the NMPC algorithm steers the post impact states to the beginning of the SS phase leading to impact invariance as suggested by Fig.~\ref{fig:ERG} (c), (f) and (i). This unusual property of the gait cycles would not be possible without the thrusters.

% Joint trajectories in a 2 by 3 grid, 1st row are angles, 2nd row are velocities

%-----------------------------------------

% DS phase GRF constraint

%-----------------------------------------
% Thrusters during DS phase

%-----------------------------------------

% ERG effect

% Use figure* to force it to span two columns

\section{Conclusion} \label{section:conclusion}

In designing closed-loop feedback for the thruster-assisted walking of bipedal robots, we assumed for well-tuned supervisory controllers and focused on fine-tuning the joint desired trajectories to satisfy the performance being sought. We devised an intermediary filter based on reference governors that guaranteed the satisfaction of performance-related constraints. We leveraged the thrusters in the system to robustify the gait cycles. Since the gait modifications and impact events can lead to deviations from the desired periodic orbits, hybrid invariance was achieved in a robust way by applying predictive schemes. The merit of the proposed approach is that unlike existing optimization-based nonlinear control methods, satisfying performance-related constraints during the single support phase does not rely on costly numeric approaches. In addition, the design allows for exploiting performance and robustness enhancing capabilities during specific parts of the gait cycle, which is unusual and not reported before.

%\clearpage
\bibliography{references}

\begin{thebibliography}{4}
\providecommand{\natexlab}[1]{#1}
\providecommand{\url}[1]{\texttt{#1}}
\providecommand{\urlprefix}{URL }
\expandafter\ifx\csname urlstyle\endcsname\relax
  \providecommand{\doi}[1]{doi:\discretionary{}{}{}#1}\else
  \providecommand{\doi}{doi:\discretionary{}{}{}\begingroup
  \urlstyle{rm}\Url}\fi

\bibitem[{Able(1956)}]{Abl:56}
Able, B. (1956).
\newblock Nucleic acid content of microscope.
\newblock \emph{Nature}, 135, 7--9.

\bibitem[{Able et~al.(1954)Able, Tagg, and Rush}]{AbTaRu:54}
Able, B., Tagg, R., and Rush, M. (1954).
\newblock Enzyme-catalyzed cellular transanimations.
\newblock In A.~Round (ed.), \emph{Advances in Enzymology}, volume~2, 125--247.
  Academic Press, New York, 3rd edition.

\bibitem[{Keohane(1958)}]{Keo:58}
Keohane, R. (1958).
\newblock \emph{Power and Interdependence: World Politics in Transitions}.
\newblock Little, Brown \& Co., Boston.

\bibitem[{Powers(1985)}]{Pow:85}
Powers, T. (1985).
\newblock Is there a way out?
\newblock \emph{Harpers}, 35--47.

\end{thebibliography}


\begin{thebibliography}{23}
\providecommand{\natexlab}[1]{#1}
\providecommand{\url}[1]{\texttt{#1}}
\providecommand{\urlprefix}{URL }
\expandafter\ifx\csname urlstyle\endcsname\relax
  \providecommand{\doi}[1]{doi:\discretionary{}{}{}#1}\else
  \providecommand{\doi}{doi:\discretionary{}{}{}\begingroup
  \urlstyle{rm}\Url}\fi

\bibitem[{Bemporad(1998)}]{bemporad1998reference}
Bemporad, A. (1998).
\newblock Reference governor for constrained nonlinear systems.
\newblock \emph{IEEE Transactions on Automatic Control}, 43(3), 415--419.

\bibitem[{{Bhat} and {Bernstein}(1998)}]{bhat1998continuous}
{Bhat}, S.P. and {Bernstein}, D.S. (1998).
\newblock Continuous finite-time stabilization of the translational and
  rotational double integrators.
\newblock \emph{IEEE Transactions on Automatic Control}, 43(5), 678--682.

\bibitem[{Chevallereau et~al.(2004)Chevallereau, Formal'sky, and
  Djoudi}]{chevallereau2004}
Chevallereau, C., Formal'sky, A., and Djoudi, D. (2004).
\newblock Tracking a joint path for the walk of an underactuated biped.
\newblock \emph{Robotica}, 22(1), 15–28.

\bibitem[{Choi and Grizzle(2005)}]{choi_grizzle_2005}
Choi, J.H. and Grizzle, J.W. (2005).
\newblock Feedback control of an underactuated planar bipedal robot with
  impulsive foot action.
\newblock \emph{Robotica}, 23(5), 567–580.

\bibitem[{{Dai} and {Tedrake}(2016)}]{7803333}
{Dai}, H. and {Tedrake}, R. (2016).
\newblock Planning robust walking motion on uneven terrain via convex
  optimization.
\newblock  \emph{IEEE-RAS International Conference on Humanoid Robots
  (Humanoids)}, 579--586.

\bibitem[{{Feng} et~al.(2014){Feng}, {Whitman}, {Xinjilefu}, and
  {Atkeson}}]{7041347}
{Feng}, S., {Whitman}, E., {Xinjilefu}, X., and {Atkeson}, C.G. (2014).
\newblock Optimization based full body control for the atlas robot.
\newblock  \emph{IEEE-RAS International Conference on Humanoid Robots},
  120--127.

\bibitem[{{Galloway} et~al.(2015){Galloway}, {Sreenath}, {Ames}, and
  {Grizzle}}]{CLFQP}
{Galloway}, K., {Sreenath}, K., {Ames}, A.D., and {Grizzle}, J.W. (2015).
\newblock Torque saturation in bipedal robotic walking through control lyapunov
  function-based quadratic programs.
\newblock \emph{IEEE Access}, 3, 323--332.

\bibitem[{Garone and Nicotra(2015)}]{garone2015explicit}
Garone, E. and Nicotra, M.M. (2015).
\newblock Explicit reference governor for constrained nonlinear systems.
\newblock \emph{IEEE Transactions on Automatic Control}, 61(5), 1379--1384.

\bibitem[{{Gilbert} et~al.(1994){Gilbert}, {Kolmanovsky}, and {Kok Tin
  Tan}}]{411031}
{Gilbert}, E.G., {Kolmanovsky}, I., and {Kok Tin Tan} (1994).
\newblock Nonlinear control of discrete-time linear systems with state and
  control constraints: a reference governor with global convergence properties.
\newblock  \emph{IEEE Conference on Decision and Control}, volume~1, 144--149.

\bibitem[{Gilbert and Kolmanovsky(2002)}]{gilbert2002nonlinear}
Gilbert, E. and Kolmanovsky, I. (2002).
\newblock Nonlinear tracking control in the presence of state and control
  constraints: a generalized reference governor.
\newblock \emph{Automatica}, 38(12), 2063--2073.

\bibitem[{{Grizzle} et~al.(2001){Grizzle}, {Abba}, and {Plestan}}]{898695}
{Grizzle}, J.W., {Abba}, G., and {Plestan}, F. (2001).
\newblock Asymptotically stable walking for biped robots: analysis via systems
  with impulse effects.
\newblock \emph{IEEE Transactions on Automatic Control}, 46(1), 51--64.

\bibitem[{{Guobiao Song} and {Zefran}(2006)}]{1641816}
{Guobiao Song} and {Zefran}, M. (2006).
\newblock Underactuated dynamic three-dimensional bipedal walking.
\newblock  \emph{IEEE International Conference on Robotics and Automation},
  854--859.

\bibitem[{Hirose and Ogawa(2006)}]{hirose2006honda}
Hirose, M. and Ogawa, K. (2006).
\newblock Honda humanoid robots development.
\newblock \emph{Philosophical Transactions of the Royal Society A:
  Mathematical, Physical and Engineering Sciences}, 365(1850), 11--19.

\bibitem[{Hurmuzlu and Marghitu(1994)}]{impact}
Hurmuzlu, Y. and Marghitu, D.B. (1994).
\newblock Rigid body collisions of planar kinematic chains with multiple
  contact points.
\newblock \emph{The International Journal of Robotics Research}, 13(1), 82--92.

\bibitem[{Khalil(2002)}]{khalil2002nonlinear}
Khalil, H. (2002).
\newblock \emph{Nonlinear Systems}.
\newblock Pearson Education. Prentice Hall.
\newblock \urlprefix\url{https://books.google.com/books?id=t\_d1QgAACAAJ}.

\bibitem[{{Kokotovic} et~al.(1992){Kokotovic}, {Krstic}, and
  {Kanellakopoulos}}]{371031}
{Kokotovic}, P.V., {Krstic}, M., and {Kanellakopoulos}, I. (1992).
\newblock Backstepping to passivity: recursive design of adaptive systems.
\newblock  \emph{IEEE Conference on Decision and Control}, volume~4,
  3276--3280.

\bibitem[{Kwon et~al.(2007)Kwon, Kim, Park, Roh, Lee, Park, Kim, and
  Roh}]{kwon2007biped}
Kwon, W., Kim, H.K., Park, J.K., Roh, C.H., Lee, J., Park, J., Kim, W.K., and
  Roh, K. (2007).
\newblock Biped humanoid robot mahru iii.
\newblock  \emph{IEEE-RAS International Conference on Humanoid Robots},
  583--588. IEEE.

\bibitem[{{Pratt} et~al.(2009){Pratt}, {Krupp}, {Ragusila}, {Rebula}, {Koolen},
  {van Nieuwenhuizen}, {Shake}, {Craig}, {Taylor}, {Watkins}, {Neuhaus},
  {Johnson}, {Shooter}, {Buffinton}, {Canas}, {Carff}, and {Howell}}]{5354430}
{Pratt}, J.E., {Krupp}, B., {Ragusila}, V., {Rebula}, J., {Koolen}, T., {van
  Nieuwenhuizen}, N., {Shake}, C., {Craig}, T., {Taylor}, J., {Watkins}, G.,
  {Neuhaus}, P., {Johnson}, M., {Shooter}, S., {Buffinton}, K., {Canas}, F.,
  {Carff}, J., and {Howell}, W. (2009).
\newblock The yobotics-ihmc lower body humanoid robot.
\newblock  \emph{IEEE/RSJ International Conference on Intelligent Robots and
  Systems}, 410--411.

\bibitem[{Raibert et~al.(2008)Raibert, Blankespoor, Nelson, and
  Playter}]{raibert2008bigdog}
Raibert, M., Blankespoor, K., Nelson, G., and Playter, R. (2008).
\newblock Bigdog, the rough-terrain quadruped robot.
\newblock \emph{IFAC Proceedings Volumes}, 41(2), 10822--10825.

\bibitem[{Raibert et~al.(1984)Raibert, Brown~Jr, and
  Chepponis}]{raibert1984experiments}
Raibert, M.H., Brown~Jr, H.B., and Chepponis, M. (1984).
\newblock Experiments in balance with a 3d one-legged hopping machine.
\newblock \emph{The International Journal of Robotics Research}, 3(2), 75--92.

\bibitem[{Sontag(1983)}]{sontag1983lyapunov}
Sontag, E.D. (1983).
\newblock A lyapunov-like characterization of asymptotic controllability.
\newblock \emph{SIAM journal on control and optimization}, 21(3), 462--471.

\bibitem[{Westervelt and Grizzle(2007)}]{westervelt2007feedback}
Westervelt, E. and Grizzle, J. (2007).
\newblock \emph{Feedback Control of Dynamic Bipedal Robot Locomotion}.
\newblock Control and Automation Series. CRC PressINC.
\newblock \urlprefix\url{https://books.google.com/books?id=xaMeAQAAIAAJ}.

\bibitem[{Westervelt et~al.(2003)Westervelt, Grizzle, and
  Koditschek}]{westervelt2003hybrid}
Westervelt, E.R., Grizzle, J.W., and Koditschek, D.E. (2003).
\newblock Hybrid zero dynamics of planar biped walkers.
\newblock \emph{IEEE transactions on automatic control}, 48(1), 42--56.

\end{thebibliography}

\end{document}